\documentclass{article}
\usepackage{spconf,amsmath,graphicx}
\usepackage{multirow}
\usepackage{graphicx}
\usepackage{enumitem}
\usepackage{xcolor}
\setlist{nosep, leftmargin=14pt}

\usepackage{mwe} 

\title{EndoOOD: Uncertainty-aware Out-of-distribution Detection in Capsule Endoscopy Diagnosis}
\name{Qiaozhi Tan$^{1,2,\dagger}$, Long Bai$^{1,\dagger}$, Guankun Wang$^{1,\dagger}$, Mobarakol Islam$^3$, Hongliang Ren$^{1,2,4,*}$\thanks{$^\dagger$ Co-first authors. $^*$ Corresponding author.}}
\address{$^1$ Department of Electronic Engineering, The Chinese University of Hong Kong, Hong Kong, China \\ $^2$ Shenzhen Research Institute, The Chinese University of Hong Kong, Shenzhen, China \\ $^3$ Wellcome/EPSRC Centre for Interventional and Surgical Sciences (WEISS) and Department of \\ Medical Physics and Biomedical Engineering, University College London, London, UK \\ $^4$ Department of Biomedical Engineering, National University of Singapore, Singapore }
\begin{document}
%
\maketitle
\begin{abstract}
Wireless capsule endoscopy (WCE) is a non-invasive diagnostic procedure that enables visualization of the gastrointestinal (GI) tract. Deep learning-based methods have shown effectiveness in disease screening using WCE data, alleviating the burden on healthcare professionals. However, existing capsule endoscopy classification methods mostly rely on pre-defined categories, making it challenging to identify and classify out-of-distribution (OOD) data, such as undefined categories or anatomical landmarks. To address this issue, we propose the Endoscopy Out-of-Distribution (EndoOOD) framework, which aims to effectively handle the OOD detection challenge in WCE diagnosis. The proposed framework focuses on improving the robustness and reliability of WCE diagnostic capabilities by incorporating uncertainty-aware mixup training and long-tailed in-distribution (ID) data calibration techniques. Additionally, virtual-logit matching is employed to accurately distinguish between OOD and ID data while minimizing information loss. To assess the performance of our proposed solution, we conduct evaluations and comparisons with 12 state-of-the-art (SOTA) methods using two publicly available datasets. The results demonstrate the effectiveness of the proposed framework in enhancing diagnostic accuracy and supporting clinical decision-making.
\end{abstract}
\section{Introduction}
\label{sec:intro}

\begin{figure}
    \centering
    \includegraphics[width=3.2in, trim=0 140 300 -20]{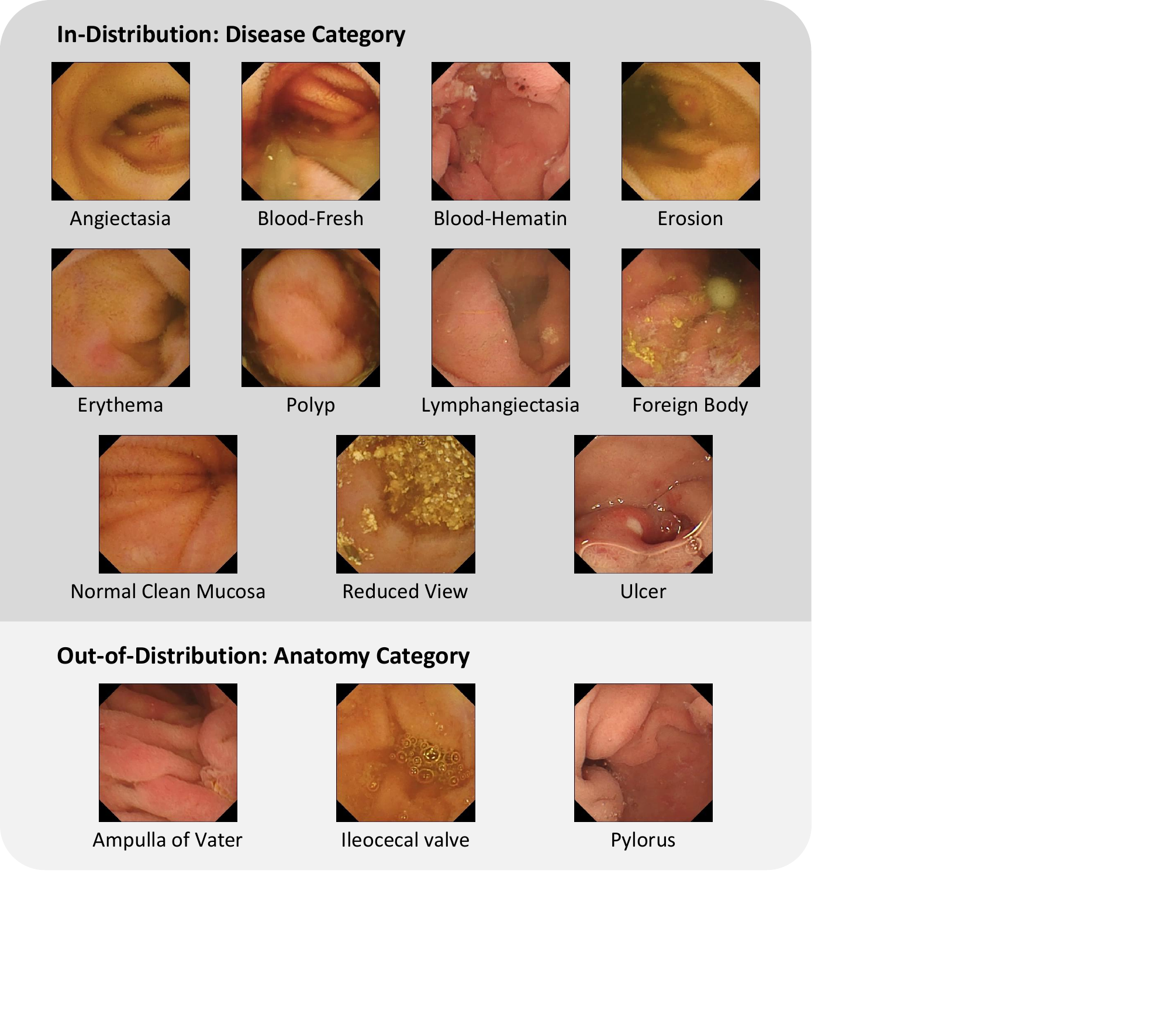}
    \caption{Image samples of the various classes for In-Distribution and Out-of-Distribution data.}
    \label{fig1}
\end{figure}

Endoscope screening has emerged as a golden standard for gastrointestinal (GI) diagnostics, enabling visualization for diagnosing and analyzing the human GI tract. Wireless capsule endoscopy (WCE) is an innovative diagnostic procedure that enables non-invasive visualization of the GI tract, offering advantages such as painlessness and the ability to reach areas inaccessible by conventional endoscopy~\cite{bai2023llcaps}. 
However, most existing classification models developed for WCE scenarios have been trained under the assumption of a closed-world setting. In this setting, data associated with pre-defined disease categories are considered as In-Distribution (ID) data. Nevertheless, the presence of Out-of-Distribution (OOD) data, such as instances from the anatomical category or anatomical landmarks interspersed within the original data, poses challenges for these models in accurately identifying and interpreting WCE images. Training models solely on ID data fail to generalize on OOD data effectively~\cite{quindos2023self}.
Fig.~\ref{fig1} shows the class distribution of two categories in the Kvasir-Capsule dataset~\cite{smedsrud2021kvasir}.
In this case, a novel framework that effectively tackles the OOD challenge can significantly enhance the robust diagnostics in WCE~\cite{yang2021generalized,wang2023rethinking,bai2023revisiting}.

To address the OOD challenge, researchers have explored diverse strategies, including training-based~\cite{devries2018learning,bai2024ossar} and post-hoc methods~\cite{wang2022vim}. MSP~\cite{hendrycks2016baseline} is the initial baseline method that relies on the maximum SoftMax score to discern between ID and OOD samples. Subsequent research efforts have delved into alternative, yet simpler and more efficient indicators for distinguishing ID and OOD instances. For instance, ODIN~\cite{liang2017enhancing} employs temperature scaling in conjunction with gradient-based input perturbations to achieve discrimination. Additionally, MDS~\cite{lee2018simple} introduces a measure based on the minimum Mahalanobis distance from class centroids, while ViM~\cite{wang2022vim} integrates the norm of feature residual against the principal space formed by training features and the original logits to compute the OOD-ness. Beyond the above methods, mixup training is becoming popular since it improves the model's ability to handle OOD samples by enhancing its predictive uncertainty estimation and enabling reliable rejection of OOD inputs~\cite{pinto2022using}. However, mixup trained models make it difficult to calibrate uncertainty, so the uncertainty estimates will be affected during training.

To solve the existing issues in WCE classification, we propose the Endoscopy Out-of-Distribution (EndoOOD) framework, comprising three key components: the uncertainty-aware mixup training strategy, the long-tailed ID data calibration, and the calibrated post-hoc inference. Our contributions to this work can be summarized as 3-fold:
\begin{itemize}
    \item [--] We propose the Endoscopy Out-of-Distribution (EndoOOD) framework. The proposed framework addresses the OOD challenges in WCE effectively by integrating uncertainty calibration solutions in the training and inference stages.
    \item [--] We incorporate the uncertainty-aware mixup training strategy to address the issue of uncertainty calibration in decoupled mixup augmentation for detecting OOD samples.  Additionally, we introduce the long-tailed ID data calibration technique to alleviate the problem of overly confident probabilities for high-frequency classes in WCE datasets.  Finally, we utilize Virtual-logit Matching (ViM) for the final calibrated inference step to distinguish between ID and OOD data.
    \item [--] Extensive experiments conducted on the Kvasir-Capsule and CIFAR-10 datasets demonstrate the superior performance of our proposed solution in addressing real-world challenges in WCE diagnosis.  This framework holds the promise of being an effective tool for gastrointestinal disease screening.
\end{itemize}

\section{Methodology}

To conduct the disease diagnosis task, we use ResNet18~\cite{he2016deep} as the backbone network, and cross entropy loss as the optimization function. During training, our method employs the uncertainty-mixup training strategy to enhance the learning process of the model. Additionally, we utilize long-tailed ID data calibration to optimize the learning objectives of the model. During inference, we introduce a ViM-based inference strategy~\cite{wang2022vim} to address information loss in the inference stage and effectively compute the presence of OOD classes.  
\subsection{Preliminaries}
\subsubsection{Mixup}
Mixup is a data augmentation technique that blends pairs of input samples and their labels by linearly interpolating them, which encourages the model to learn from the combined information of different examples~\cite{zhang2018mixup}. For a simple $\left(x, y\right)$, the mixup sample is defined as:
\begin{equation}
    x_{mixup}=\alpha x + (1-\alpha)x^{'} 
\end{equation}
where $\alpha \in (0,1)$, $x_{mixup}$ consists entirely of $x^{'}$ when $\alpha$ is 0. Mixup can help improve OOD detection by increasing the model's ability to generalize to unseen data.  By blending samples and labels, Mixup introduces a form of regularization that encourages the model to learn more robust and discriminative features, which can help the model better differentiate between ID and OOD samples~\cite{ravikumar2023intra,pinto2022using}.

\subsubsection{Virtual-logit Matching}
Virtual-logit Matching (ViM)~\cite{wang2022vim} defines a principal subspace using the eigenvectors of the largest eigenvalues of the training data covariance matrix. The residual of a test sample is extracted as the projection of its features onto the orthogonal complement space of the principal subspace, and the virtual logits are obtained by the rescaled residual. Then, the logits are added to the original logits of the inference samples and propagated through the SoftMax function to achieve the ViM scores. A higher ViM score indicates a higher likelihood of the sample being OOD. By incorporating both the feature residual space and the original logit space, ViM achieves improved OOD detection performance compared to methods that rely on only one source of information.

\subsection{Uncertainty-Aware Mixup Training}
In mixup training for OOD detection, uncertainty plays a crucial role in measuring the model's confidence and identifying OOD inputs.
Although mixup training offers a way to enhance the model's ability to distinguish between in-distribution and OOD inputs, it can lead to decreased calibration performance, impacting the accurate estimation of uncertainty. To overcome this calibration degradation, encouraged by~\cite{wang2023pitfall}, we follow a decoupling principle that separates mixup's data transformation and random perturbation steps. The mixup inference process is translated into the training. Specifically, in every block of the ResNet18, we recover the prediction of $x$, which adopts the same coefficient $\alpha$ as in the training phase. 

In this case, raw sample output can be approximately recovered. Then, models can be trained by fitting the decoupled outputs to the original one-hot labels, thereby mitigating the effects of the confidence penalty resulting from label smoothing. Besides, we incorporate a large $margin$ between $\alpha_1$ and $\alpha_2$ during data transformation, which can effectively reduce the impact of noise. This can be achieved by sampling $\alpha_1$ and $\alpha_2$ from the intervals [0.5, 1] and [0, 0.5], respectively, or by imposing a constraint that requires both $\alpha_1$ and $\alpha_2$ to be greater than a specific constant. In this process, our models can be trained from the original one-hot labels while avoiding the calibration issues typically associated with mixup. This strategy solves the calibration problem and improves the predictive performance compared to vanilla mixup.

\subsection{Long-Tailed ID Data Calibration}
Due to the large variation in the incidence of different symptoms, capsule endoscopy diagnosis tends to have long-tailed data, which refers to a distribution of data where a few categories have a high frequency. ID Long-tailed data can cause the model to produce lower probabilities for low-frequency categories, thus making it unable to distinguish them correctly from OOD samples. Conventional model calibration techniques such as temperature scaling (TS) and label smoothing (LS) often struggle to handle long-tail data effectively, as they tend to mitigate the issue of overconfident predictions through the manipulation of either one-hot labels or predicted probabilities. This manipulation can lead to poor calibration of the model's predictions and affect its overall performance.

To address the above challenge, inspired by~\cite{islam2021class}, we first incorporate category quantity with TS. By reweighting the optimal temperature based on category quantity, the model can adjust its confidence scores based on the distribution of classes, mitigating overconfidence in high-frequency classes. For max-normalized category quantity $\left[q_{1}, q_{2}, \ldots q_{N}\right]$ with $N$ total classes, our modified temperature is defined as follows:
\begin{equation}
    T^{cq}=T^{o}+\beta\left[q_{1}, q_{2}, \ldots, q_{N}\right]
\end{equation}
where $T^{o}$ is optimal temperature and $\beta$ is a hyperparameter set to 0.1 in our experiments. Then, the category quantity optimal temperature vector is integrated with distillation loss, further reducing self-distillation miscalibration.

Furthermore, we introduce category quantity into LS. By penalizing high-frequency classes more heavily, the modified LS ensures better calibration across the entire dataset. It is formulated as:
\begin{equation}
    s^{cq}=s^{o}+\gamma\left[q_{1}, q_{2}, \ldots, q_{N}\right]
\end{equation}
where $\gamma$ is also a hyperparameter and set to 0.01.

\begin{table*}[t]
\caption{OOD detection results on Kvasir-Capsule and CIFAR-10 datasets. We use FPR@95, AUROC, AUPR In and AUPR Out as the metric for OOD detection. LTDC denotes long-tailed ID data calibration.}
\label{tab:main} 
\resizebox{\linewidth}{!}{
\begin{tabular}{c|ccccccccc}
\hline
\multirow{2}{*}{Methods} & \multicolumn{4}{c|}{Near-OOD (Kvasir-Capsule outliers)} & \multicolumn{4}{c|}{Far-OOD (CIFAR-10)} & Kvasir-Capsule inliers   \\\cline{2-10} 
                  & \multicolumn{1}{l}{FPR@95$\downarrow$} & \multicolumn{1}{l}{AUROC$\uparrow$} & \multicolumn{1}{l}{AUPR In$\uparrow$} & \multicolumn{1}{l|}{AUPR Out$\uparrow$}      & \multicolumn{1}{l}{FPR@95$\downarrow$} & \multicolumn{1}{l}{AUROC$\uparrow$} & \multicolumn{1}{l}{AUPR In$\uparrow$} & \multicolumn{1}{l|}{AUPR Out$\uparrow$}       & Acc$\uparrow$                                 \\ \hline
ConfBranch~\cite{devries2018learning}     & 78.59                       & 69.25                      & 58.39                        & \multicolumn{1}{c|}{76.97}          & 82.47                       & 74.00                      & 94.71                        & \multicolumn{1}{c|}{26.92}           & 92.32                                     \\
G-ODIN~\cite{hsu2020generalized}         & 62.24                       & 80.64                      & 73.00                        & \multicolumn{1}{c|}{85.47}          & 6.74                        & 99.11                      & 99.88                        & \multicolumn{1}{c|}{95.14}           & 92.34                                     \\
MSP~\cite{hendrycks2016baseline}            & 84.49                       & 76.84                      & 69.54                        & \multicolumn{1}{c|}{79.10}          & 43.29                       & 87.93                      & 97.71                        & \multicolumn{1}{c|}{62.40}           & 92.85                                     \\
ODIN~\cite{liang2017enhancing}           & 78.22                       & 80.39                      & 74.58                        & \multicolumn{1}{c|}{80.88}          & 9.71                        & 97.77  & 99.66                        & \multicolumn{1}{c|}{88.11}           & 92.85                                     \\
MDS~\cite{lee2018simple}            & 49.80                       & 88.10                      & 85.13                        & \multicolumn{1}{c|}{91.00}          & \textbf{0.00}                        & \textbf{100.00}                     & \textbf{100.00}                       & \multicolumn{1}{c|}{99.98}           & 92.85                                     \\
MDSENS~\cite{lee2018simple}         & 82.84                       & 75.90                      & 72.69                        & \multicolumn{1}{c|}{78.46}          & 0.22                        & 99.90                      & 99.99                        & \multicolumn{1}{c|}{99.35}           & 92.85                                     \\
RMDS~\cite{ren2021simple}           & 94.02                       & 80.83                      & 76.91                        & \multicolumn{1}{c|}{77.68}          & 6.50                        & 98.32                      & 99.75                        & \multicolumn{1}{c|}{87.58}           & 92.85                                     \\
REACT~\cite{sun2021react}         & 67.67                       & 77.78                      & 65.07                        & \multicolumn{1}{c|}{83.60}          & 37.08                       & 89.21                      & 97.78                        & \multicolumn{1}{c|}{69.31}           & 92.85                                     \\
VIM~\cite{wang2022vim}           & 39.61                       & 92.49                      & 90.26                        & \multicolumn{1}{c|}{94.22}          & \textbf{0.00}                        & \textbf{100.00}                     & \textbf{100.00}                       & \multicolumn{1}{c|}{99.99}           & 92.85                                     \\
KNN~\cite{sun2022out}           & 70.99                       & 85.06                      & 81.72                        & \multicolumn{1}{c|}{87.43}          & 6.04                        & 98.81                      & 99.80                        & \multicolumn{1}{c|}{95.54}           & 92.85                                     \\
DICE~\cite{sun2022dice}          & 83.82                       & 77.27                      & 72.77                        & \multicolumn{1}{c|}{77.46}          & 33.43                       & 93.62                      & 99.05                        & \multicolumn{1}{c|}{64.74}           & 92.85                                     \\
ASH~\cite{djurisic2022extremely}           & 68.00                       & 76.70                      & 66.86                        & \multicolumn{1}{c|}{82.91}          & 38.50                       & 87.05                      & 97.16                        & \multicolumn{1}{c|}{69.95}           & 92.85                                     \\
 \hline
Ours w/o LTDC        & 34.70                       & 93.03                      & 90.48                        & \multicolumn{1}{c|}{\textbf{94.80}} & \textbf{0.00}                        & \textbf{100.00}                     & \textbf{100.00}                       & \multicolumn{1}{c|}{99.98}           & 93.52                                     \\
Ours w/i LTDC (T=0.01)    & 37.59        & 92.70       & 90.77        & \multicolumn{1}{c|}{94.43}              & \textbf{0.00}        & \textbf{100.00}       & \textbf{100.00}         & \multicolumn{1}{c|}{\textbf{100.00}}                & 93.58                      \\
Ours w/i LTDC (T=0.05)    & \textbf{32.08}              & \textbf{93.06}             & \textbf{90.92}               & \multicolumn{1}{c|}{93.75}          & \textbf{0.00}               & \textbf{100.00}            & \textbf{100.00}              & \multicolumn{1}{c|}{99.98}           & \textbf{94.02}                            \\
Ours w/i LTDC (T=0.10)    & 33.82        & 92.72       & 90.57         & \multicolumn{1}{c|}{93.83}             & \textbf{0.00}        & \textbf{100.00}       & \textbf{100.00}        & \multicolumn{1}{c|}{\textbf{100.00}}                & 93.93                      \\ \hline
\end{tabular}
}
\end{table*}

\begin{table}[t]
\caption{Ablation on the weight parameters of the uncertainty-aware mixup training. The bolded weights are the parameters we adopted in the above experiments. UAMT denotes uncertainty-aware mixup training.}
\label{tab:abl} 
\resizebox{\linewidth}{!}{
\begin{tabular}{c|c|ccccc}
\hline
\multirow{2}{*}{\textcolor{white}{aa}$\alpha$\textcolor{white}{aa}} & \multirow{2}{*}{margin} & \multicolumn{4}{|c|}{Near-OOD (Kvasir-Capsule Outliers)} & Kvasir-Capsule Inliers   \\\cline{3-7} 
    &     & FPR@95$\downarrow$        & AUROC$\uparrow$         & AUPR In$\uparrow$        & \multicolumn{1}{c|}{AUPR Out$\uparrow$} & Acc$\uparrow$ \\ \hline
\multicolumn{1}{c|}{0.1}       & \multicolumn{1}{c|}{0.0}       & 34.86               & 92.67               & 89.97               & \multicolumn{1}{c|}{94.74}               & 92.79                      \\
\multicolumn{1}{c|}{0.5}       & \multicolumn{1}{c|}{0.0}       & 42.31               & 92.23               & 90.09               & \multicolumn{1}{c|}{93.96}                              & 93.48                      \\
\multicolumn{1}{c|}{1.0}       & \multicolumn{1}{c|}{0.0}       & 44.85               & 89.61               & 86.66               & \multicolumn{1}{c|}{91.03}               & 93.47                      \\
\multicolumn{1}{c|}{\textbf{0.1}} & \multicolumn{1}{c|}{\textbf{0.5}} & \textbf{32.08} & \textbf{93.06} & \textbf{90.92} & \multicolumn{1}{c|}{93.75}  & \textbf{94.02}        \\
\multicolumn{1}{c|}{0.5}       & \multicolumn{1}{c|}{0.5}       & 34.10               & 92.61               & 89.61               & \multicolumn{1}{c|}{\textbf{94.75}}                              & 94.02             \\
\multicolumn{1}{c|}{1.0}       & \multicolumn{1}{c|}{0.5}       & 37.17               & 92.07               & 89.42               & \multicolumn{1}{c|}{93.87}                            & 93.93                      \\ \hline
\multicolumn{2}{c|}{w/o UAMT}  & 35.29               & 91.97               & 88.72              & \multicolumn{1}{c|}{94.32}                            & 92.85\\\hline
\end{tabular}}
\end{table}

\section{Experiments}
\subsection{Dataset}
We evaluate our method on Kvasir-Capsule~\cite{smedsrud2021kvasir}. and CIFAR-10~\cite{krizhevsky2009learning} datasets. Kvasir-Capsule~\cite{smedsrud2021kvasir} is a large video capsule endoscopy (VCE) dataset that consists of 47,238 frames labeled data with 11 disease classes and 3 anatomical classes. We set the disease classes as the ID data, and the anatomical classes as the Near-OOD data in our experiments, respectively. Furthermore, to comprehensively evaluate the model's OOD detection performance, we also employ the CIFAR-10 dataset~\cite{krizhevsky2009learning} as the Far-OOD data, which is a widely used benchmark consisting of 60,000 color images of 10 different object classes. CIFAR-10 and WCE datasets exhibit significant differences in appearance, making CIFAR-10 a suitable benchmark for evaluating the Far-OOD performance.

\subsection{Implementation Details}
We compare our proposed solution against several SOTA methods, including ConfBranch~\cite{devries2018learning}, G-ODIN~\cite{hsu2020generalized}, MSP~\cite{hendrycks2016baseline}, ODIN~\cite{liang2017enhancing}, MDS~\cite{lee2018simple}, MDSENS~\cite{lee2018simple}, RMDS~\cite{ren2021simple}, REACT~\cite{sun2021react}, VIM~\cite{wang2022vim}, 
KNN~\cite{sun2022out}, DICE~\cite{sun2022dice}, 
and ASH~\cite{djurisic2022extremely}. 
We train our models using the SGD optimizer for $100$ epochs with a batch size of $128$. The learning rate is $1 \times 10^{-1}$ and weight decay is $1 \times 10^{-5}$. All experiments are conducted using NVIDIA A100 and the Python PyTorch framework.

\subsection{Experimental Results}

We perform a quantitative evaluation against SOTA methods, as shown in Table~\ref{tab:main}. Our proposed methodology demonstrates exceptional performance across accuracy and OOD metrics, surpassing all baseline models. Specifically, for the Near-OOD data, our method achieves 32.08, 93.06, 90.92, and 93.75 in FPR@95, AUROC, AUPR In, and AUPR Out, respectively. This indicates the enhanced discriminative ability of our methods in distinguishing unknown classes. Moreover, our methods demonstrate improved recognition capabilities for Far-OOD data, as evidenced by the error-free performance in the CIFAR-10 dataset. Notably, although the observed improvement in unknown classes, the accuracy of known classes still increases by more than 1\%. Through appropriately calibrating the uncertainty of mixup training and overly confident predictions, our EndoOOD framework effectively cultivates an excellent OOD detection model.

\subsection{Ablation Study}

In Table~\ref{tab:main} and Table~\ref{tab:abl}, we assess the effects of removing the (i) Uncertainty-aware Mixup Training and (ii) Long-Tailed ID Data Calibration. Experimental results reveal that the absence of any of these methods leads to a significant degradation in model performance. These observations indicate that our proposed methods positively contribute to achieving the best results. Besides, we assess the effects of different category quantity temperature values in Table~\ref{tab:main}. Experimental results reveal that setting T=0.05 yields the best performance.

Furthermore, Table~\ref{tab:abl} investigates the impact of different mixup training weights $\alpha$ and margin ($\alpha$ differences between two mixing samples) on the model's performance through grid search. We end up using the setting $\alpha$=0.1 and margin=0.5, as it achieves the best performance.

\section{Conclusion}
In this paper, we present the EndoOOD framework for uncertainty-aware OOD detection in capsule endoscopy diagnosis. The framework utilizes a combination of techniques, including uncertainty-aware mixup training, long-tailed in-distribution (ID) data calibration, and ViM-based calibrated post-hoc inference, to address the challenges associated with OOD data in WCE scenarios.  Experimental results on the Kvasir-Capsule and CIFAR-10 datasets have shown that the proposed framework outperforms existing methods in terms of OOD detection metrics. The framework enhances the model's ability to handle OOD samples, improves uncertainty estimation, and enables reliable rejection of OOD inputs, thereby improving the overall diagnostic accuracy and clinical decision-making in WCE.  Future research can explore further advancements and applications of the EndoOOD framework in real-world WCE scenarios to benefit patients and healthcare professionals.

\section{Acknowledgments}
\label{sec:acknowledgments}
This work was supported by the Hong Kong Research Grants Council (RGC) CRF C4063-18G, GRF 14203323, GRF 14216022, GRF 14211420, NSFC/RGC Joint Research Scheme Grant N\_CUHK420/22; Shenzhen-Hong Kong-Macau Technology Research Programme (Type C) STIC Grant SGDX20210823103535014 (202108233000303).


\bibliographystyle{IEEEbib}
\bibliography{strings,refs}

\end{document}